\def\year{2022}\relax
\documentclass[letterpaper]{article} % DO NOT CHANGE THIS
\usepackage{aaai22}  % DO NOT CHANGE THIS
\usepackage{times}  % DO NOT CHANGE THIS
\usepackage{helvet}  % DO NOT CHANGE THIS
\usepackage{courier}  % DO NOT CHANGE THIS
\usepackage[hyphens]{url}  % DO NOT CHANGE THIS
\usepackage{graphicx} % DO NOT CHANGE THIS
\usepackage{natbib}  % DO NOT CHANGE THIS AND DO NOT ADD ANY OPTIONS TO IT
\usepackage{caption} % DO NOT CHANGE THIS AND DO NOT ADD ANY OPTIONS TO IT
\DeclareCaptionStyle{ruled}{labelfont=normalfont,labelsep=colon,strut=off} % DO NOT CHANGE THIS
\usepackage{booktabs}
\usepackage{algorithmic}
\usepackage{adjustbox}
\usepackage{multirow}
\usepackage{subfigure} 
\usepackage{bbm}
\usepackage{float}
\usepackage{changes}
\usepackage{amsmath}
\usepackage{amssymb}
\usepackage[ruled,vlined]{algorithm2e}

\SetKwInput{KwInput}{Input}                % Set the Input
\SetKwInput{KwOutput}{Output}              % set the Output
\SetKwInput{KwInit}{Initialization}         % initialization 

\usepackage{newfloat}
\usepackage{listings}
\floatstyle{ruled}
\newfloat{listing}{tb}{lst}{}
\floatname{listing}{Listing}

\title{Delving into Sample Loss Curve to Embrace Noisy and Imbalanced Data}
\author {
    Shenwang Jiang\textsuperscript{\rm 1,2}
    Jianan Li\textsuperscript{\rm 1,2,*}
    Ying Wang\textsuperscript{\rm 1,2}
    Bo Huang\textsuperscript{\rm 1,2}
    Zhang Zhang \textsuperscript{\rm 3}
    Tingfa Xu \textsuperscript{\rm 1,2,}\thanks{Corresponding author}
}
\affiliations {
    \textsuperscript{\rm 1} School of Optics and Photonics, Image Engineering \& Video Technology Lab, Beijing Institute of Technology\\
    \textsuperscript{\rm 2} Key Laboratory of Photoelectronic Imaging Technology and System, Ministry of Education of China\\
    \textsuperscript{\rm 3} Department of Computer Science, University of Massachusetts Lowell \\
     \{jiangwenj02, lijianan15, wangying7275\}@gmail.com, a1039377853@163.com \\ Zhang\_Zhang@student.uml.edu, ciom\_xtf1@bit.edu.cn
}
\usepackage{bibentry}
\newcommand{\bd}[1]{\textbf{#1}}

\begin{document}

\maketitle

\begin{abstract}
Corrupted labels and class imbalance are commonly encountered in practically collected training data, which easily leads to over-fitting of deep neural networks (DNNs).  Existing approaches alleviate these issues  by adopting a sample re-weighting strategy, which is to re-weight sample by designing weighting function. However, it is only applicable for training data containing only either one type of data biases.
In practice, however, biased samples with corrupted labels and of tailed classes commonly co-exist in training data.
How to handle them simultaneously is a key but under-explored problem. In this paper, we find that these two types of biased samples, though have similar transient loss, have distinguishable trend and characteristics in loss curves, which could provide valuable priors for sample weight assignment. Motivated by this, we delve into the loss curves and propose a novel probe-and-allocate training strategy: In the probing stage, we train the network on the whole biased training data without intervention, and record the loss curve of each sample as an additional attribute; In the allocating stage, we feed the resulting attribute to a newly designed curve-perception network, named CurveNet, to learn to identify the bias type of each sample and assign proper weights through meta-learning adaptively. 
The training speed of meta learning also blocks its application.
To solve it, we propose a method named skip layer meta optimization (SLMO)  to accelerate training speed by skipping the bottom layers.
Extensive synthetic and real experiments well validate the proposed method, which achieves state-of-the-art performance on multiple challenging benchmarks.
Code is available at \url{https://github.com/jiangwenj02/CurveNet-V1}.
\end{abstract}

\noindent {Deep neural networks (DNNs)~\cite{hu2018squeeze, simonyan2014very, he2016deep} have made tremendous progress thanks to the rapid growth of labeled training data \cite{5206848,10.1007/978-3-319-10602-1_48}. However, practically collected training samples always suffer from corrupted labels~\cite{zhang2016understanding} and class imbalance~\cite{he2009learning}, which easily causes over-fitting of DNNs and leads to poor generalization capability. This robust deep learning issue has attracted increasing attention recently. ~\cite{zhang2017range,liu2019large,kang2019decoupling,shu2019meta,tan2020equalization}.}

{Sample re-weighting approach is a commonly adopted strategy to mitigate the above robust learning issue, which aims to learn a weighting function mapping training loss to sample weight. However, such re-weighting strategy fails to handle training data with both corrupted labels and class imbalance, since there exist two entirely contradictive requirements for constructing the loss-weight mapping for the two types of biased data. Specifically, for training data containing corrupted labels, samples with corrupted labels tend to have large training loss, so the weighting function is supposed to map large loss to small sample weight to mitigate the effect of label noise. In contrast, for training data with class imbalance, samples of tailed classes usually suffer large loss due to insufficient training, so the weighting function ought to assign large weights to these hard positive samples, making the network emphasize more on the tailed classes to improve overall performance. Given the fact that most practically collected data suffer both corrupted labels and class imbalance, solving the two types of data bias simultaneously is a challenging and under-explored task~\cite{cao2020heteroskedastic}.}

%-------------------------------------------------------------------------

\begin{figure}[t]
\begin{center}
   \includegraphics[width=1.0\linewidth]{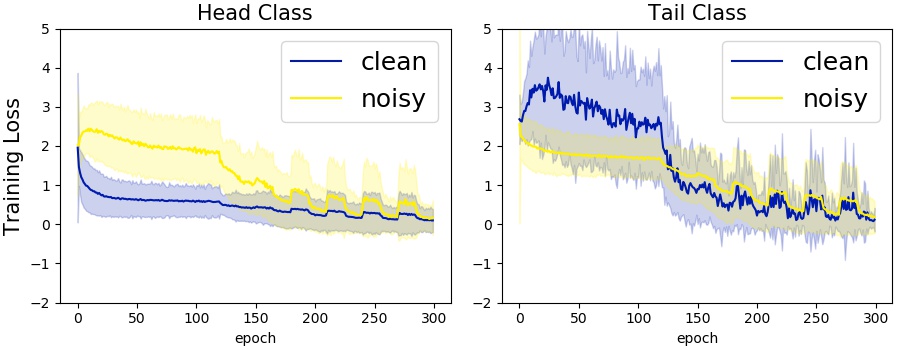}
\end{center}
\vspace{-0.60cm} 
   \caption{
   % The normalized loss value curves of different classes. The curve represents the average value of all samples in this class, and the shaded areas indicate the respective variances.
   Average training loss along with variance of clean and noisy samples. Though noisy and clean samples become indistinguishable from transient loss over training, the trend of their loss curves are dramatically different in the long run. Such difference could provide valuable priors to distinguish the two types of biased data and assign proper sample weights accordingly through meta-learning, making it feasible to embrace biased training data with both corrupted labels and class imbalance for model training.}
\label{fig:costcurve}
\vspace{-0.60cm} 
\end{figure}

%-------------------------------------------------------------------------

A key challenge is to distinguish clean samples of tail class from those with corrupted labels, i.e., noisy samples, and to assign different weights accordingly. Figure~\ref{fig:costcurve} illustrates training loss of samples of head and tail class. The blue and the yellow curve represent the average training loss along with variance for clean and noisy samples, respectively. For head class, noisy samples have larger loss than clean ones throughout the training, making them easy to be distinguished. For tail class, however, it is non-trivial to distinguish them simply by transient loss since the loss values of clean and noisy samples become very close over training. By going deeper into the loss curve, we found that noisy and clean samples demonstrate obviously different loss trend in the long run. Concretely, the loss of noisy samples remains stable at the beginning of the training, while the loss of clean samples rises sharply in the beginning and then falls quickly. Hence, the training loss curve in fact encodes valuable information and could provide useful priors to distinguish clean and noisy samples of tailed classes.

In light of this, we propose to take advantage of the informative training loss curve to distinguish clean samples of tail class from noisy samples, and generate proper sample weights accordingly. To this end, we propose a novel probe-and-allocate training strategy: In the probing stage, we train a classifier with cyclical learning rate on the entire biased training data, and record the loss curve of each sample. To highlight the loss difference between clean and noisy samples, we normalize the loss of each sample by subtracting the mean loss of samples of the same class and dividing by the standard deviation; In the allocating stage, we take the normalized loss curve as an attribute of each sample, and further attach embedded class label to facilitate the identification of noise. We feed the loss curve along with the class embedded label to a newly designed curve-perception network, named CurveNet, to capture the overall characteristics of the loss curve and output a dynamic corresponded weight further referred by loss function. Inspired by Meta-Weight-Net~\cite{shu2019meta}, we adopt meta-learning to optimize the allocating stage to produce large and small weights for clean samples of tail class and noisy samples, respectively, thus making the classifier emphasis more on the hard positive samples while being robust to noise.
It is well known that training speed has caused a real bottleneck in current meta-learning methods.
To solve it, we propose a method called SLMO to skip the bottom layers in classifier when we train the CurveNet, which can save a lot of calculations in backward while maintaining the performance of the CurveNet.

We comprehensively evaluate the proposed approach on a series of biased training datasets by manually adjusting noise and imbalance ratios. Thanks to the informative loss curve prior collected in the probing stage, the newly designed CurveNet can distinguish different types of biased sample and assign proper sample weights accordingly in the allocating stage. The resulting classifier can be well optimized using biased training data with both corrupted labels and class imbalance, achieving state-of-the-art performance on CIFAR10, CIFAR100 and Clothing1M.

To sum up, the main contributions of this work are:
\begin{itemize}
\item We propose a novel probe-and-allocate training strategy, paving a new way for embracing biased training data with both corrupted labels and class imbalance.
\item A new CurveNet is designed to exploit informative loss curve to distinguish different types of biased data and generate proper sample weight accordingly. 
\item SLMO is proposed to speed up the training speed of meta learning approaches and maintain the training effect. 
\item The proposed method establishes new state-of-the-arts on multiple datasets, and is also generic and extendable for training models with noisy and imbalanced data on various recognition tasks. 
\end{itemize}

%------------------------------------------------------------------------
\section{Related Works}

Most previous efforts~\cite{kumar2010self,pi2016self,hendrycks2018using,ma2018dimensionality} focus on solving either corrupted labels or class imbalance.
Some methods like Meta-Weight-Net \cite{shu2019meta} can alleviate these two problems separately, but still fail to solve both problems together.
Therefore, we introduce the methods related to these two issues.

\noindent\bd{Corrupted labels.}
Methods to address the corrupted labels pay more attention to easy samples with smaller losses, such as self-paced learning series \cite{kumar2010self,jiang2014easy,jiang2014self} and curriculum learning \cite{bengio2009curriculum}.
The SPL series simultaneously selects easy samples from all the samples and learns from the new easy samples \cite{pi2016self}.
Another popular approach attempts to introduce noise-robust loss functions like the ramp loss \cite{brooks2011support}, the unhinged loss \cite{van2015learning} and the savage loss \cite{masnadi2008design}, which is robust against corrupted labels.
Some popular approaches attempt at correcting corrupted labels by a supplemental clean label inference step, such as GLC \cite{hendrycks2018using}, Reed \cite{reed2014training}, Co-training \cite{han2018co}, D2L\cite{ma2018dimensionality}, and S-Model \cite{goldberger2016training}.
O2U-Net \cite{huang2019o2u} considers samples with higher average loss have a higher probability of being noisy labels.
Based on it, O2U-Net filters noisy samples by the average loss.

\noindent\bd{Class imbalance.}
The methods to solve the imbalance of classes are mainly divided into two categories.
One is to weight for each class according to its frequency, and the other is to weight for each sample according to its loss value. 
The pioneer of the former methods weights by the inverse of the class frequency \cite{huang2016learning,wang2017learning} or the inverse square root \cite{mahajan2018exploring,mikolov2013distributed}.
The latter methods aim to study the training difficulty of  samples in terms of their loss and assign higher weights to hard training samples, such as \cite{freund1997decision,lin2017focal,malisiewicz2011ensemble,dong2017class}.
There are also methods based on transfer learning \cite{wang2017learning,cui2018large} that transfer the knowledge of classes with a large amount of samples to the classes with less samples.

\noindent\bd{Meta Learning.}
Meta-learning \cite{finn2017model,antoniou2018train,li2017meta,shu2018small,ravi2016optimization} is a method of optimizing the network with the second-order derivative.
Typical methods based on meta-learning include L2T-DLF \cite{wu2018learning}, MentorNet \cite{jiang2018mentornet}, L2RW \cite{ren2018learning}, and Meta-Weight-Net \cite{shu2019meta}.
L2T-DLF is composed of the teacher model and the student model.
The teacher model plays the role of outputting loss functions to train the student model.
MentorNet aims to supervise the training of the base deep networks and to generate a suitable weight for the current sample by a bidirectional LSTM network.
Inspired by L2RW and Meta-Weight-Net, our method uses the same meta-learning training strategy.
The difference is that a fixed attribute for each sample is used to generate a suitable weight, but the input of L2RW and Meta-Weight-Net varies with training and cannot represent the overall training state of the sample. 
This attribute can indicate the training difficulty of the sample and whether it is noise. 
Therefore, our method can solve class imbalance and corrupted labels in the meantime.

%------------------------------------------------------------------------

%------------------------------------------------------------------------

\section{Our Method}
\subsection{Revisiting Meta-Weight-Net}
\label{sec:revisit}

\begin{figure*}[t]
\begin{center}
   \includegraphics[width=0.85\linewidth]{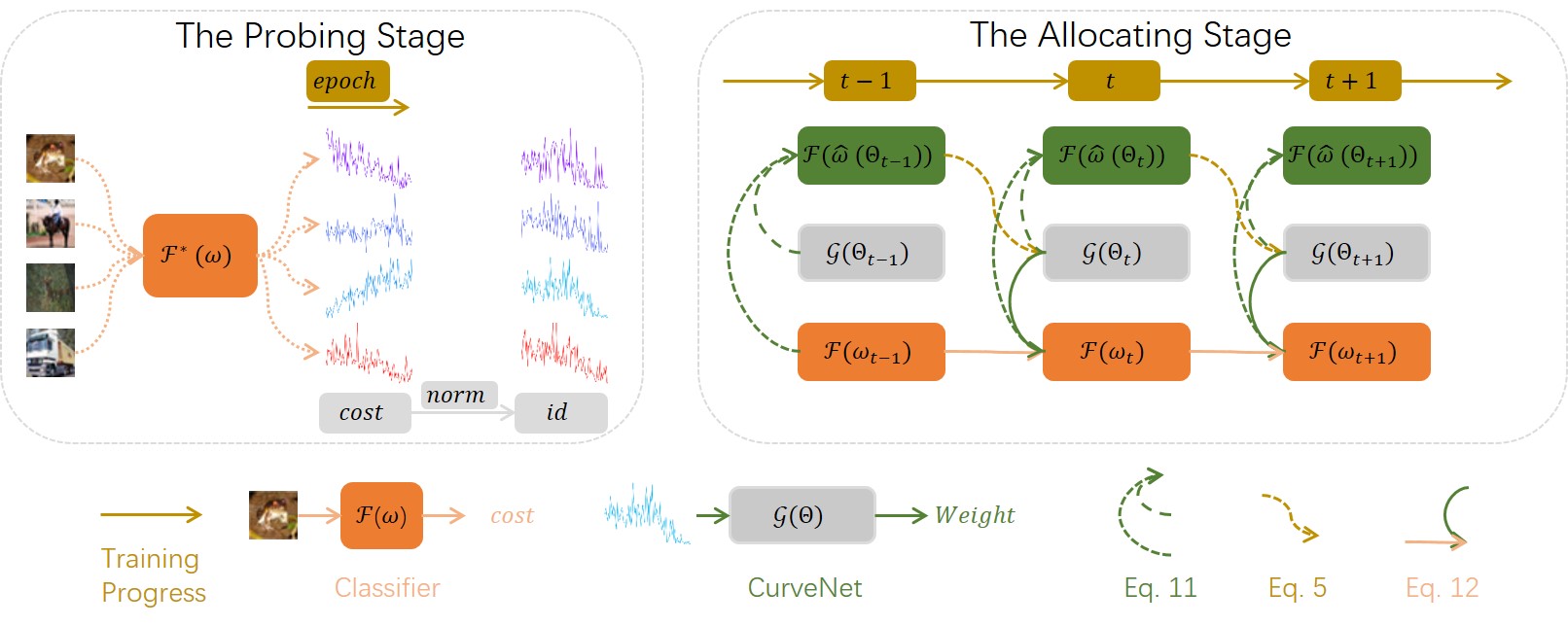}
\end{center}
\vspace{-0.30cm} 
   \caption{Overall workflow of the proposed probe-and-allocate training strategy: the probing stage trains a classifier on the entire biased dataset to collect training loss curve for each sample; the allocating stage first re-weights the loss curves by integrating the loss curve and class embedding through a newly designed CurveNet, and then generates parameters of the classifier for different types of biased data through meta-learning.
   }
\label{fig:framework}
\vspace{-0.50cm} 
\end{figure*}

Meta-Weight-Net is proposed to learn a classification network $\mathcal{F}$ effectively with biased training data. 
Building on the idea of meta-learning, a small additional unbiased meta data set (with clean labels and balanced data distribution) is employed to refine the parameters $\omega$ of the classifier $\mathcal{F}$.
For notation convenience, we denote the biased training-data as $\mathcal{D}^{tra}=\{x^{tra}_i,y^{tra}_i\}_{i=1}^N$ while the unbiased meta-data as $\mathcal{D}^{meta}=\{x^{meta}_i,y^{meta}_i\}_{i=1}^M$,  where $N$ and $M$ respectively cope to sample's amount and $N \gg M$.
We denote $X^{tra}$,$Y^{tra}$ as the set of all training data and label of $\mathcal{D}^{tra}$ respectively.
$X^{meta}$ and $Y^{meta}$ are defined by the same way.

For traditional training methods, the parameters of the classifier can be obtained by minimizing the loss function in the following form:
\vspace{-0.1cm} 
\begin{equation}
    \omega^* = \mathop{\arg\min}_\omega \mathcal{L}(Y^{tra},\mathcal{F}(X^{tra}|w)),
    \label{eq:eq1}
\vspace{-0.15cm} 
\end{equation}
where $\mathcal{F}$ always acts as a convolutional neural network.
In the following we denote $\mathcal{L}_{tra}$ as $\mathcal{L}(Y^{tra},\mathcal{F}(X^{tra}|w))$.
However, given the existence of the biased data, the training process tends to be sub-optimal easily if it follows the previous way.
% re-weighting baunch
Aiming to enhance the robustness of training, a re-weighting method is adopted to impose weight $\mathcal{G}(\mathcal{L}_{tra}|\Theta)$ on the sample loss, where $\mathcal{G}$ represents the weight net and $\Theta$ represents the parameters of $\mathcal{G}$. 
The final loss is expressed as the weighted sum of the weight net $\mathcal{G}$ and the original loss.
% 一旦 thata 定了，w就能取到最优值
Once $\Theta$ is set, the optimal value $\omega^*$ can be therefore determined.
Thus, the Equation~\ref{eq:eq1} can be further formed as:
\vspace{-0.1cm} 
\begin{equation}
    \omega^*  = \mathop{\arg\min}_\omega  \mathcal{G}(\mathcal{L}_{tra} |\Theta)\mathcal{L}_{tra}.
\vspace{-0.15cm} 
\end{equation}

Specifically, $\mathcal{G}$ is composed of an MLP network with only one hidden layer, containing 100 nodes, and using Sigmoid as the activation function. To guarantee the output within the interval of $[0,1]$, the sigmoid activation function is adopted after the output layer. 
The parameters $\Theta$ are optimized through meta-learning methods, which minimizes the loss function applied on the meta-data set mentioned above as:
\begin{equation}
    \Theta^* = \mathop{\arg\min}_\Theta\mathcal{L}(Y^{meta},\mathcal{F}(X^{meta} |\omega^*(\mathcal{G}(\Theta)))).
    \label{eq:theta}
\end{equation}
In the following we denote this loss function by $\mathcal{L}_{meta}$.
Since the two parameters $w$ and $\Theta$ need to be optimized at the same time, a separate optimization method is used by first treating $\Theta$ as a known quantity and finding the optimal solution of $w$ on a mini-batch presented by Equation~\ref{eq:fl}, which is used to continue optimizing $\Theta$ through Equation~\ref{eq:ug}. 
Here $t$ represents the current training epoch.
\begin{equation}
    \hat{\omega}^{t} = \omega^{t} - \alpha \bigtriangledown_\omega  \mathcal{G}(\mathcal{L}^{t}_{tra}|\Theta^{t}) \circ  \mathcal{L}^{t}_{tra}|_{\omega^{t}},
    \label{eq:fl}
\end{equation}
where $\circ$ denotes element-wise product.
Subsequently, $\Theta$ and $w$ can be calculated through:
\begin{equation}
    \Theta^{t+1} = \Theta^{t} - \beta \bigtriangledown_\Theta \mathcal{L}^{t}_{meta}(\hat{\omega}^{t}(\Theta^{t})) |_{\Theta^{t}}.
    \label{eq:ug}
\end{equation}
\begin{equation}
    \omega^{t+1} = \omega^{t} - \alpha \bigtriangledown_\omega \mathcal{G}(\mathcal{L}^{t}_{tra}|\Theta^{t+1}) \circ  \mathcal{L}^{t}_{tra}|_{\omega^{t}}.
    \label{eq:uf}
\end{equation}

Despite the superior results achieved by Meta-Weight-Net, it has inherent disadvantages to be solved.
First, the meta sub-network adopts the current loss value as input which changes dramatically throughout the training procedure and fails to represent the sample's state. 
Second, the loss value varies at each epoch and gets smaller and smaller within the training process, which is not conducive to network convergence. 
Moreover, when noise and hard samples present at the same time, the weights could be either large or small, resulting in unsatisfactory performance of the classifier. 
For the purpose of issue-solving, we propose a novel probe-and-allocate strategy and adopt a newly designed CurveNet for loss refinement. More details will be explained in the next section.

\subsection{Overall Structure} \label{sec:os}
Our network is structured on the basis of Meta-Weight-Net\cite{shu2019meta} with modified loss value and weight network $\mathcal{G}$. As shown in Figure \ref{fig:framework}, the whole structure is composed of two stages of probing-stage as main-network for classification and allocating-stage for parameters refinement. 

\begin{figure}[t]
\begin{center}
   \includegraphics[width=1.0\linewidth]{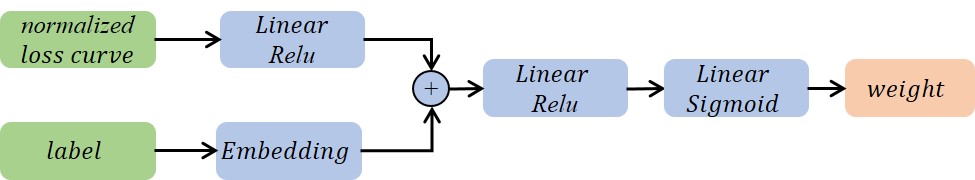}
\end{center}
\vspace{-0.30cm} 
   \caption{The network architecture of Curvenet, which takes normalized loss curve and class label as input and outputs a proper weight for each sample adaptively.
   }
\label{fig:CurveNet}
\vspace{-0.70cm} 
\end{figure}

In the probing-stage, the biased training data is fed into the classifier and the loss values are obtained by implementing a loss function between the predict label and the ground-truth. 
As a loss value is difficult to present the full picture of the sample, we delve into the loss curves throughout the whole training process and find distinguishable trends and characteristics between the noisy sample and clean tail sample. 
From Figure~\ref{fig:costcurve}, it can be seen that the loss of the noisy samples stabilizes at a value when the learning rate is at a high value, while that of the clean samples rises sharply at the beginning and then decreases quickly.
Inspired by this, we innovatively propose CurveNet(as shown in Figure ~\ref{fig:CurveNet}) to replace the MLP structure used in Meta-Weight-Net\cite{shu2019meta}, fully leveraging the loss curve information to adjust and integrate the loss values. 

Similar to the re-weighting network mentioned in Meta-weight-net, the allocating-stage adopts the meta-learning idea and allows the weighted loss values to guide the training of the classification network, giving the classifier more emphasis on hard positive samples while being robust to noise. 
This training approach well feeds the loss information to the training process of the classification network's parameters, which is very effective for parameters refinement. 
A detailed description of the CurveNet is given in the next section.

\subsection{CurveNet}

Based on the analysis above, we argue that the whole loss curve is far more informative than a single value, which ought to be utilized as a whole. 
Gathering all of the loss value $l_{i,t}$ of the $i$th sample together as a one-dimensional vector $L_i=[l_{i,0}, l_{i,1}, \cdots, l_{i,T}]$, where $T$ represents the number of training epoch. 
As the parameters of the classifier are randomly initialized, the loss values of the previous epoch vary irregularly and thus have no reference value. 
Consequently, the first S loss values are removed from the one-dimensional vector, so that the loss vector of the $i$th sample can be expressed as: $L_i=[l_{i,S}, l_{i,S+1}, \cdots, l_{i,T}]$. 

We then normalize the loss of each sample through subtracting the average loss of samples in the same category to highlight the loss distinction between clean and noisy samples:
\vspace{-0.20cm} 
\begin{equation}
  \mu_{{k,t}} = \frac{\sum_j^N \mathbbm{1}(k,y_j)l_{j,t}}{\sum_j^N \mathbbm{1}(k,y_j)},
  \label{eq:mean}
\end{equation}
% \begin{equation}
%     \sigma_{t} = \frac{1}{N-1}\sum_{j=1}^{N}{(l_{j,t} - \mu_{{t}})^2}
%     \label{eq:std}
% \end{equation}
\vspace{-0.20cm} 
\begin{equation}
    % \hat{l}_{i,t} = \frac{l_{i,t} - \mu_{t}}{\sigma_{t}} 
    \overline{l}_{i,t} = l_{i,t} - \mu_{y_i,t}.
    \label{eq:norm}
\end{equation}
% \vspace{-0.1cm} 
Here we denotes K as the number of class ($1 \le k \le K$) and $\mathbbm{1}$ as a Dirac delta function.
$\mathbbm{1}(k,y_j)$ equals 1 when $k$ equals $y_j$, otherwise 0.

The normalized loss vectors can be denoted as $I$, which are then fed sequentially into fully connected layers, each coupled to a ReLU activation layer. $P$ is the number of output neurons of the last fully connected layer, which is set as $64$ here through experiments.

As a way to further facilitate noise identification, we adopt the class label embedding method to enrich the class information into the loss curve features. Such embedding method is commonly used in the field of natural language processing \cite{cao2020heteroskedastic}, and the embedded matrix here could be expressed as $Y^{K \times P} = [Y_1,\cdots,Y_K]$. 
Then, the summation result of loss curve feature and label embedded feature is performed as the input of two sequential fully connected layers, each of which is appended with the active functions of ReLU and Sigmiod, respectively. As mentioned above, the sigmoid function ensures all weights fall within an interval.

CurveNet performs as a change-sensitive network that is responsible for capturing trends in loss values, successfully distinguishing different types of biased samples, and assigning appropriate sample weights accordingly in the assignment phase.

\subsection{Skip Layer Meta Optimization}
Currently, slow training blocks the advancement of meta-learning methods, and most of the time is consumed in Equation \ref{eq:theta} for optimizing $\Theta$.
According to FaMUS \cite{Xu2021FaMUS}, the Equation $\bigtriangledown_\Theta \mathcal{L}^{t}_{meta}|_{\Theta^{t}}$ can be rewritten by the chain rule as follows, 
\begin{equation}
\begin{split}
     \bigtriangledown_\Theta \mathcal{L}^{t}_{meta}|_{\Theta^{t}}  & = \frac{\partial \mathcal{L}^{t}_{meta} }{\partial \hat{\omega}^t} \bullet \frac{\partial \hat{\omega}^t}{\partial \mathcal{G}(\Theta^{t})} \bullet \frac{\partial \mathcal{G}(\Theta^{t})}{\partial \Theta^{t}} \\
     & \varpropto \sum_i^Z \frac{\partial \mathcal{L}^{t}_{meta} }{\partial \hat{\omega}^t_i} \bullet \frac{\partial \hat{\omega}^t_i}{\partial \mathcal{G}(\Theta^{t})} \bullet \frac{\partial \mathcal{G}(\Theta^{t})}{\partial \Theta^{t}},
\end{split}
    \label{eq:ug_s}
\end{equation}
where, $Z$ represents the number of layers in classifer.
From the Equation \ref{eq:ug_s}, we know that the amount of computation to optimize $\Theta$ is positively related to $Z$.
Based on it, we propose skip layer meta optimization (SLMO), which freezes the bottom layers when we optimize the $\Theta$.  
SLMO can be formulated as follows,
\begin{equation}
     \bigtriangledown_\Theta \mathcal{L}^{t}_{meta}|_{\Theta^{t}}   \varpropto \sum_{i=SL}^Z \frac{\partial \mathcal{L}^{t}_{meta} }{\partial \hat{\omega}^t_i} \bullet \frac{\partial \hat{\omega}^t_i}{\partial \mathcal{G}(\Theta^{t})} \bullet \frac{\partial \mathcal{G}(\Theta^{t})}{\partial \Theta^{t}},
    \label{eq:ug_sK}
\end{equation}
where, $SL$ is the number of frozen layers.

\subsection{Training Method}

Considering the input of CurveNet is the loss curve of the samples, the Equation \ref{eq:fl} and \ref{eq:uf} should be modified as follows:
\vspace{-0.20cm} 
\begin{equation}
    \hat{\omega}^{t} = \omega^{t} - \alpha \bigtriangledown_\omega \mathcal{G}([I, Y^{tra}]|\Theta^{t}) \circ \mathcal{L}^{t}_{tra}|_{\omega^{t}},
    \label{eq:fl_our}
\end{equation}
\vspace{-0.20cm} 
\begin{equation}
    \omega^{t+1} = \omega^{t} - \bigtriangledown_\omega \alpha \mathcal{G}([I, Y^{tra}]|\Theta^{t+1}) \circ \mathcal{L}^{t}_{tra}|_{\omega^{t}}.
    \label{eq:uf_our}
\end{equation}

It is worth noting that when the learning rate is changed, there are significant differences in the loss value curves for different classes of samples.
Therefore, cyclical learning rate \cite{smith2017cyclical} is adopted to train the classifier $\mathcal{F}(\omega)$ in the probing stage, which is also employed by O2U-Net \cite{huang2019o2u}.
Besides, when the learning rate of the classifier decreases, we consider that the CurveNet has been optimized well and do not update the parameters of the CurveNet anymore to speed up the training.

%-------------------------------------------------------------------------
\section{Experiments}
We use CIFAR dataset~\cite{krizhevsky2009learning} with varying noise rates and imbalance ratios to verify the effectiveness of the propose method. We also test our method on real noisy and imbalanced data to validate its generality. 

\subsection{Datasets} \label{sec:dataset}

\noindent\bd{CIFAR-10}. 
This dataset consists of 60,000 RGB images (50,000 for training and 10,000 for testing). Images are equally distributed to 10 categories.
We randomly select 100 images from each category to form unbiased meta data set.

\begin{figure}[t]
\centering
\subfigure{
\begin{minipage}[t]{0.48\linewidth}
\centering
\includegraphics[width=0.9\linewidth]{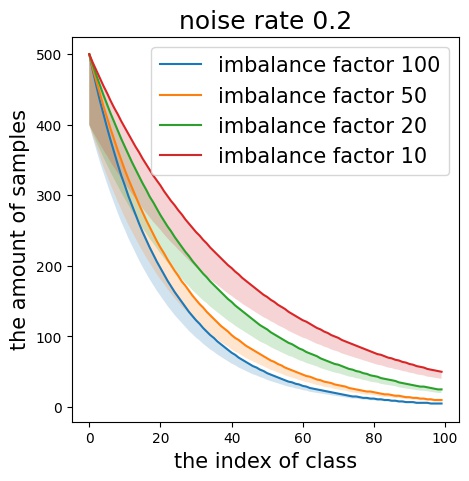}
\end{minipage}%
}%
\subfigure{
\begin{minipage}[t]{0.48\linewidth}
\centering
\includegraphics[width=0.9\linewidth]{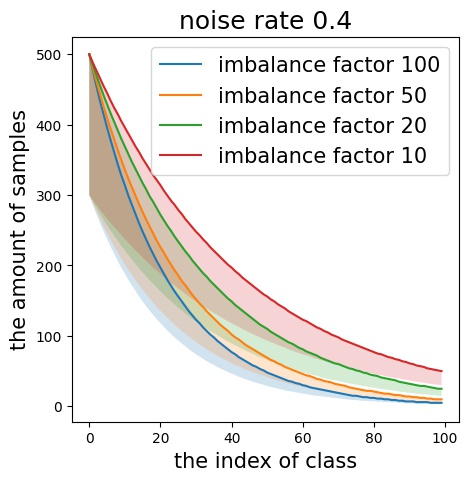}
\end{minipage}
}%
\centering
\vspace{-0.30cm} 
\caption{
The number of all samples (solid line) and noisy samples (shadow below) of each class in CIFAR-100 with varying imbalance factors and noise rates.
}
\label{fig:imb_noise_0}
\vspace{-0.50cm} 
\end{figure}

\noindent\bd{CIFAR-100}. This dataset comprises 100 categories, each of which contains 600 images.
We randomly select 10 images from each category to form our unbiased meta-data set.
Figure~\ref{fig:imb_noise_0} shows the varied sample number for each class when adjusting the imbalance factor and noise rate. 

\noindent\bd{Clothing1M}. This dataset~\cite{xiao2015learning} comprises 1M clothing images of 14 categories crawled from online shopping websites. 
The image labels are mainly generated from surrounding text provided by sellers, leading to many corrupted labels and a certain imbalance. 
The dataset also provides additional verified clean data for training.

\noindent\bd{Food-101N}. This dataset~\cite{lee2018cleannet} is a large-scale dataset (310k/25k training/test images) accompanied by 55k images with clean verification labels.

\noindent\bd{Class imbalance}. We gradually reduce the number of samples in each class using the exponential function $n_i=n_0\mu^i$, where $n_i$ is  the sample number of class $i$ and $\mu \in (0,1]$.
We use class imbalance factor to measure how imbalanced the data is, which is defined as the sample number of the most frequent (head) class divided by that of the least frequent (tail) class. 
The first column in Figure~\ref{fig:imb_noise_0} shows the sample number of each class in imbalanced CIFAR-100 with imbalance factor ranging from 10 to 200.

\noindent\bd{Corrupted labels}. 
Commonly adopted label noise types include uniform noise and flipping noise, which randomly corrupt the label of a sample from its true class to any other class and a specified class, respectively, with a fixed probability of $p$ (noise rate). 
In this work, we follow the Meta-Weight-Net and adopt flip2 noise with noise rate $p$ on CIFAR10, which randomly corrupts true labels with probability $p$ to two other random classes. 

%-------------------------------------------------------------------------
\subsection{Implementation Details} \label{sec:implementaion}

\noindent\bd{CIFAR}. We construct biased training dataset with varying noisy and imbalance ratios by manually adjusting the sample number of each class and adding corrupted labels to the clean and balanced dataset such as CIFAR10 and CIFAR100.
{To explore the effect of our method on more scenarios, we conduct experiments on cifar10 with various imbalance ratios and rates of flip2 noise , which emphasizes on imbalance ratios.
Meanwhile, studies on cifar100 with various imbalance ratios and rates of uniform noise  are carried out, which emphasizes on noise rates.}
We train the model for 200 epochs on a single NVIDIA GTX 1080Ti. 
During allocating stage, we use stochastic gradient descent (SGD) with initial learning rate 0.1 and decrease the learning rate to 0.01 and 0.001 at epoch 80 and 100, respectively. 
We use a batchsize of 128 images. CurveNet is optimized using Adam with learning rate 0.001.
We choose 10 and 100 samples from each category to form the meta data set of cifar100 and cifar10 respectively. 

\noindent\bd{Clothing1M and Food101-N}. 
We use ResNet50 as the classifier, and adopt SGD as the optimizer and step learning schedule to optimize it.

\subsection{Comparison Methods.} 
We select three types of methods for comparison: 1) methods to solve class imbalance, such as Focal Loss~\cite{lin2017focal}, Class-Balanced~\cite{cui2019class},  and LDAM-DRW~\cite{cao2019learning}; 2) methods to solve corrupted label, such as Co-teaching~\cite{han2018co}, O2UNet~\cite{huang2019o2u}, Bootstrapping~\cite{reed2014training}, S-adaptation~\cite{goldberger2016training}, LCCN~\cite{li2019learning}, CleanNet~\cite{lee2018cleannet}, MetaCleaner~\cite{zhang2019metacleaner}, Self-Learning~\cite{han2019deep}, and Distill~\cite{zhang2020distilling}; and 3) methods to solve class imbalance and corrupted label simultaneously such as Meta-weight-Net~\cite{shu2019meta} and HAR \cite{cao2020heteroskedastic}.

\begin{table}[t]
\begin{center}
\begin{adjustbox}{max width=0.4\textwidth}
\begin{tabular}{c|c|c}
\hline
Dataset & CIFAR10 & CIFAR100 \\
\hline
Imbalance ratio & [1,10,20,50,100,200] & [1,10,20] \\
\hline
Noise rate & [0.0,0.2,0.4] & [0.0,0.2,0.4,0.6] \\
\hline
\hline
CE Loss  & 74.49 & 46.76 \\
\hline
\hline
Class-Balanced & 63.49 & 42.81 \\
\hline
Focal & 71.59 & 43.85 \\
\hline
LDAM-DRW & 73.46 & 45.47  \\
\hline
\hline
Co-teaching & 60.63 & 36.55   \\
\hline
O2U & 65.01 & 40.21  \\
\hline
\hline
MW-Net(noise) & 74.13 & 49.28  \\
MW-Net(imb) & 71.54 & 49.20 \\
HAR& 73.50 & 42.88 \\
\hline
\hline
Our & \bd{75.70} & \bd{50.49}  \\
\hline
\end{tabular}
\end{adjustbox}
\vspace{-0.3cm}
\caption{Performance comparisons on CIFAR10 and CIFAR100 with varying noise rates and imbalance factors.
The best results are highlighted in \bd{bold}.}
\label{tab:cifar10_res}
\end{center}
\vspace{-0.5cm}
\end{table}
%-------------------------------------------------------------------------

\subsection{Image Classification on CIFAR10} \label{sec:vis_cifar10}

We conduct extensive studies on 18($6 \times 3$) setting biased CIFAR10 with varying imbalance factors [1, 10, 20, 50, 100, 200] and noise rates [0, 0.2, 0.4].
Table~\ref{tab:cifar10_res} presents the average accuracy of different methods using ResNet-32 as backbone.
We propose a new metric, that is, the mean accuracy (MA) under all imbalance factors and noise rates to give a comprehensive comparison of models in handling different biased data. 
In this metric, our method achieves the best performance.
Interestingly, the CE Loss model achieves higher MA than many carefully designed methods due to the fact that such methods are specialized for specific biased data while MA cares more about overall performance.

\begin{table}[t]
\begin{center}
\begin{adjustbox}{max width=0.47\textwidth}
    \begin{tabular}{c|c|c|c|c|c|c}
    \hline
    Imbalance ratio & 50    & 100   & 200   & \multicolumn{3}{c}{1} \\
    \hline
    Noise rate & \multicolumn{3}{c|}{0} & 0     & 0.2   & 0.4 \\
    \hline
    MW-Net(imb) & 77.55 & 72.38 & 63.08 & 91.78 & 90.29 & 87.51 \\
    MW-Net(noise) & 72.12  & 64.57  & 58.34  & 92.99  & 90.80  & 88.25  \\
    Ours  & \textbf{78.71} & \textbf{73.52} & \textbf{65.91} & \textbf{93.23} & \textbf{92.01} & \textbf{90.69} \\
    \hline
    \end{tabular}%
\end{adjustbox}
\vspace{-0.3cm}
\caption{Performance comparisons on CIFAR10 with varying noise rates or imbalance factors.
}
\label{tab:cifar10_res_d}
\vspace{-0.8cm}
\end{center}
\end{table}

Table \ref{tab:cifar10_res_d} reports the result of our method and MW-Net on biased CIFAR10 with either corrupted label or class imbalance.
When the noise rate equals 0, it becomes a pure class imbalance task. 
Our method outperforms MW-Net(imb) by a large margin, i.e., $1.16\%$, $1.14\%$, and $2.83\%$ when the imbalance factor equals 50, 100 and 200 respectively.
When the imbalance factor equals 1, it becomes a pure corrupted label task. 
Compared with MW-Net(noise), our method boosts the accuracy by $1.21\%$ and $2.44\%$ at the noise rate of $20\%$ and $40\%$, respectively.
One can see that our method gets the best performance in accuracy, even when the noise rate equals 0 and the imbalance factor equals 1.

\begin{figure*}[h]
\begin{center}
   \includegraphics[width=0.95\linewidth]{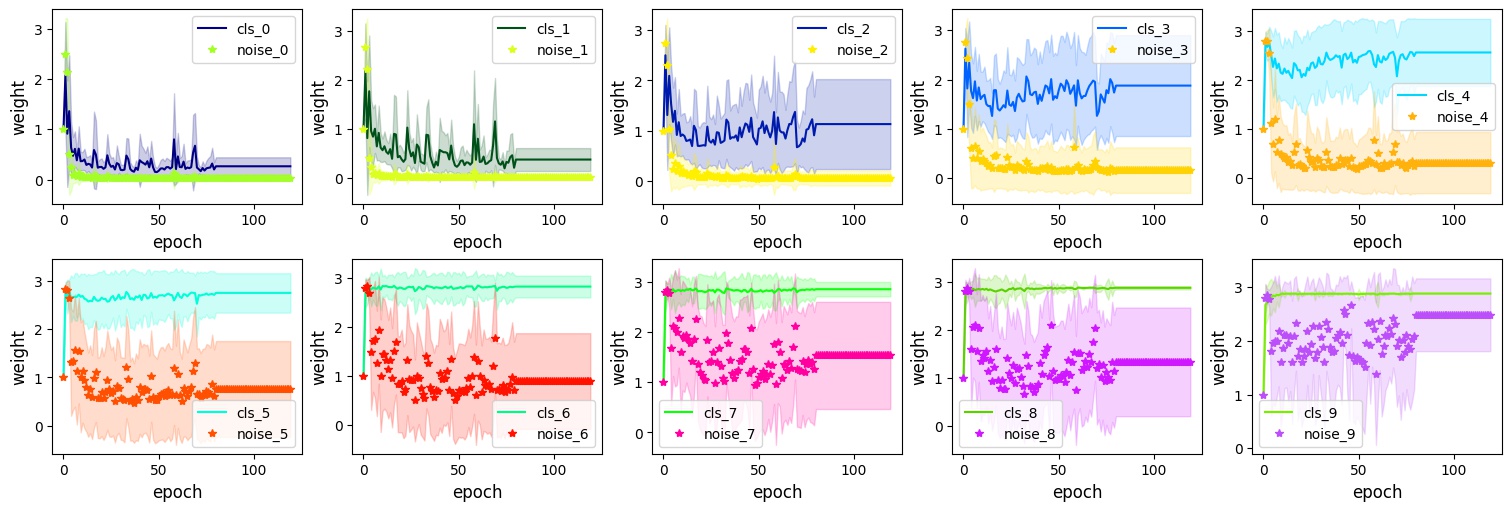}
\end{center}
\vspace{-0.50cm} 
   \caption{
   The weight of clean and noisy samples of all classes on CIFAR-10 with imbalance factor 20 and noise rate 0.4.}
\label{fig:side:weightinclass}
\vspace{-0.50cm} 
\end{figure*}

\noindent\bd{Qualitative Analyses.} 
We demonstrate the weights for clean and noisy samples of all the classes in Figure \ref{fig:side:weightinclass}.
It is clearly observable that in all the classes with different amount of samples our method distinguishes noisy and clean samples well and gives a larger weight to the class with less samples.

\begin{figure}[t]
\begin{center}
   \includegraphics[width=1.0\linewidth]{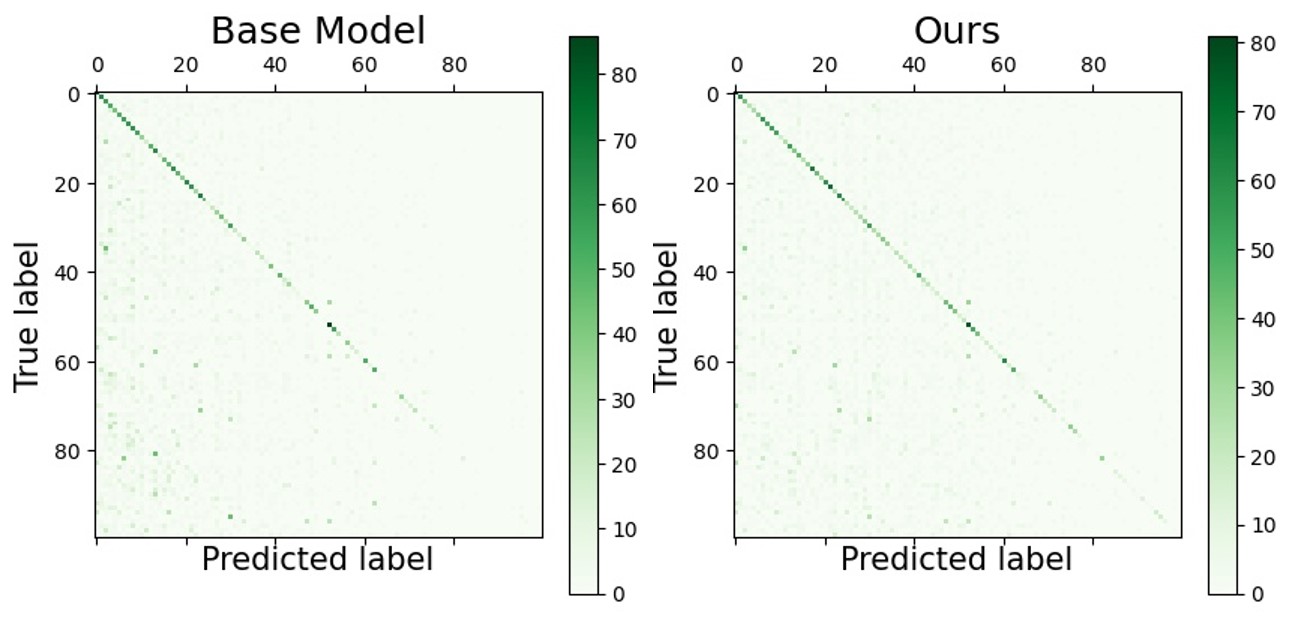}
\end{center}
\vspace{-0.50cm} 
   \caption{
   Confusion matrices for the CE Loss model and our model on CIFAR-100 with imbalance factor 20 and noise rate 0.6.}
\label{fig:side:confusion}
\vspace{-0.00cm} 
\end{figure}

%------------------------------------------------------------------------

\begin{table}[t]
\begin{center}
\begin{adjustbox}{max width=0.47\textwidth}
    \begin{tabular}{c|c|c|c|c|c|c}
    \hline
    Imbalance ratio &  \multicolumn{3}{c|}{10}    & \multicolumn{3}{c}{20} \\
    \hline
    Noise rate & 0.2   & 0.4   & \multicolumn{1}{c|}{0.6} & 0.2   & 0.4   & \multicolumn{1}{c}{0.6} \\
    \hline
    % basemodel & 70.09  & 57.57  & 52.00  & 58.57  & 46.33  & 88.45  \\
    % \hline
    MW-Net(imb) & 49.71  & \textbf{44.08 } & 30.68  & 42.82  & 34.90  & 22.35  \\
    \hline
    MW-Net(noise) & 51.12  & 42.17  & 30.34  & \textbf{44.32}  & 36.33  & \textbf{25.53 } \\
    % \hline
    % focal loss & 70.54  & 55.70  & 50.72  & 65.36  & 60.47  & 51.47  \\
    \hline
    Ours  & \textbf{51.93 } & 43.97  & \textbf{31.94 } & 44.29  & \textbf{37.88 } & 25.07  \\
    \hline
    \end{tabular}%
\end{adjustbox}
\vspace{-0.3cm}
\caption{Performance comparisons on CIFAR100 with varying noise rates or imbalance factors.
}
\label{tab:cifar100_res_d}
\vspace{-0.8cm}
\end{center}
\end{table}

%-------------------------------------------------------------------------
\subsection{Image Classification on CIFAR100}
We also test the proposed method on 12($3 \times 4$) setting biased CIFAR100 with various imbalance factors [1, 10, 20] and noise rates [0.0, 0.2, 0.4, 0.6]. 
As depicted in Table~\ref{tab:cifar10_res}, our method outperforms MW-Net(noise) by $1.21\%$ in MA, achieving the best performance among most priors with carefully designed loss functions.
Table \ref{tab:cifar100_res_d} reports the result of our method and MW-Net on biased CIFAR100 with both corrupted label and class imbalance.
With imbalance factor 20 and noise rate 0.4, our method gets a remarkable boost in accuracy (2.98\%) over MW-Net(imb).
Our method achieves the best performance under almost all settings of imbalance factors and noise rates.
The confusion matrices of CIFAR-100 with imbalance factor 20 and noise rate 0.6 are displayed in Figure~\ref{fig:side:confusion}.
One can see that our method can effectively improve the accuracy for tail classes by greatly reducing incorrect classification.

\subsection{Results on Large-scale Dataset.}
\noindent\bd{Clothing1M.}
For a fair comparison with previous work~\cite{shu2019meta}, we use ResNet-50 as backbone in this experiment.
The setting of CurveNet is the same as that on Cifar.
Table~\ref{tab:clothing} shows the proposed method outperforms MW-Net by 0.69\% in accuracy and achieves the best performance compared to other priors, evidencing the superiority of the proposed method in handling real biased data.

\begin{table}[h]
\vspace{-0.3cm}
  \centering
  \begin{adjustbox}{max width=0.47\textwidth}
    \begin{tabular}{c|cccccc}
    \toprule
    Method & CE Loss & Bootstrapping & S-adaptation & LCCN & MW-Net & Ours \\
    \midrule
    Acc.(\%)  & 68.94 & 68.94 & 68.94 & 73.07 & 73.72 & \bd{74.41} \\
    \bottomrule
    \end{tabular}%
    \end{adjustbox}
    \vspace{-0.4cm}
    \caption{Performance comparisons on Clothing1M.}
    \label{tab:clothing}
    \vspace{-0.4cm}
\end{table}%

\noindent\bd{Food-101N.}
We further evaluate our method on Food-101N.
For fairness, we compare with preeminent methods that also use clean data. Table~\ref{tab:food101} shows our method performs on par with SOTAs on Food-101N, evidencing its superior generalization capability on large-scale datasets.
\begin{table}[h]
\vspace{-0.3cm}
  \centering
  \begin{adjustbox}{max width=0.47\textwidth}
    \begin{tabular}{c|cccccc}
    \toprule
    Method & CE Loss & CleanNet & MetaCleaner & Self-Learning & Distill & Ours \\
    \midrule
    Acc.(\%)  & 81.44 & 83.96 & 85.05 & 85.11 & \bd{87.57} & 87.19 \\
    \bottomrule
    \end{tabular}%
    \end{adjustbox}
    \vspace{-0.4cm}
    \caption{Performance comparisons on Food-101N.}
    \vspace{-0.4cm}
  \label{tab:food101}%
\end{table}%

\subsection{Ablation Study}

We conduct ablation study on CIFAR10 with various imbalance factors [1, 10, 20] and noise rates [0.0, 0.2, 0.4, 0.6], and set the noise type as uniform to further validate our approach can handle different noise.

\noindent
\bd{Universality and Scalability.} 
To verify the universality of our method on different backbones, besides ResNet-34, we further test with other popular backbones, i.e., WRN-16-8, on CIFAR10.
Table~\ref{tab:cifar10_wrn} shows our method with WRN-16-8
also performs well under varying noise rates and imbalance ratios.
Furthermore, consistent accuracy improvement can be achieved by further integrating DRW strategy. 
This well evidences the scalability of our approach.
\begin{table}[h]
\vspace{-0.3cm}
\begin{center}
\begin{adjustbox}{max width=0.47\textwidth}
    \begin{tabular}{c|c|c|c|c|c|c}
    \hline
    Imbalance ratio & \multicolumn{3}{c|}{10}    & \multicolumn{3}{c}{20} \\
    \hline
    Noise rate  & 0.2   & 0.4 & 0.6  & 0.2   & 0.4 & 0.6 \\
    \hline
    CE Loss & 78.81 & 66.11 & 57.11 & 73.46 & 64.68 & 45.17 \\
    \hline
    MW-Net (noise) & 75.66 & 70.57 & 58.71 &   78.26    & 60.95 & 46.09 \\
    \hline
    Ours  & 85.94 & 80.51 & 73.55 & \bd{83.84} & 76.95 & 65.77 \\
    \hline
    Ours+DRW & \bd{86.52} & \bd{81.24} & \bd{74.92} & 83.74 & \bd{77.28} & \bd{66.78} \\
    \hline
    \end{tabular}%
\end{adjustbox}
\vspace{-0.3cm}
\caption{
Test accuracy of our model with  WRN-16-8 and DRW on CIFAR-10.
}
\label{tab:cifar10_wrn}
\vspace{-0.7cm}
\end{center}
\end{table}

\noindent\bd{Meta Data.} 
Considering the difficulty in collecting unbiased data, we further test the robustness of our approach by reducing the meta data used for training.
Table~\ref{tab:cifar10_size} shows our method can still brings 1.24\% boost in accuracy on CIFAR10 (noise rate 20\% and imbalance ratio 20) even when only a  minimal number of 10 unbiased images are used. The improvement becomes larger as the used unbiased images increase.
Besides, we also try a new strategy to construct meta dataset without using extra unbiased data.
Theoretically, training samples with lower loss in the probing stage are more likely to be clean data. 
By selecting 10 low-loss samples (per category) as unbiased meta dataset in the allocating stage, 
our approach still brings 1.13\% improvement,
suggesting its strong robustness to varying amount and quality of unbiased data. 

\begin{table}[h!]
\vspace{-0.3cm}
\begin{center}
\begin{adjustbox}{max width=0.47\textwidth}
    \begin{tabular}{c|c|c|c|c|c|c|c|c}
    \toprule
    Method  & CE     & \multicolumn{6}{c|}{Ours} & MWNet\\
    \hline
    Num.  & 100  & 0   & 10    & 20    & 50    & 80    & 100 & 100\\
    \hline
    Acc.(\%)  & 77.89 & \bd{\underline{\textit{79.02}}} & 79.13     & 79.21 & 79.51 & 80.46 & \bd{81.61} & 78.81\\
    \bottomrule
    \end{tabular}%
\end{adjustbox}
\vspace{-0.3cm}
\caption{
Test accuracy of our model with different sacles of meta data on CIFAR-10.
}
\label{tab:cifar10_size}
\vspace{-0.4cm}
\end{center}
\end{table}

\noindent\bd{Effects of Skip-Layer.} Table \ref{tab:cifar10_skiplayer} analyzes the impact of the number of skip-layer on the CIFAR10 data set with various noise rates and imbalance factors. 
As shown in the table, the training time decreases significantly with the increase of skip-layer yet the accuracy is slightly reduced.
For example, compared with SL=0, the speed boosts by 5.71 times and the MA only decreases by 0.93\% when SL=3.

\begin{table}[h]
% \vspace{-0.1cm}
\begin{center}
\tabcolsep=0.30cm
\begin{tabular}{c|c|c|c|c}
\hline
 &  SL=0  & SL=1 & SL=2  &  SL=3 \\
 \hline
 MA (\%)  &  \bd{82.46} & 81.99 & 81.81 & 81.53 \\
 \hline
  time(ms) & 239.94 & 222.45 & 65.44  & 41.98 \\
\hline
\end{tabular}%
\vspace{-0.30cm} 
\caption{
    Test accuracy and training time of our method with different skip layers on CIFAR-10. }
\label{tab:cifar10_skiplayer}
\vspace{-0.3cm}
\end{center}
\end{table}

% \FloatBarrier

\noindent\bd{Effects of Embedding Dimension.} We test our model with different embedding dimension P on the synthetic Cifar10, and the results are putted in Table \ref{tab:cifar10_P}.
It can be seen that the MA has a slight increase as P increases.
Considering the training time and the performance, we prefer to set P equal to 64.  
\begin{table}[h]
\vspace{-0.1cm}
\begin{center}
\tabcolsep=0.37cm
\begin{tabular}{c|c|c|c|c}
\hline
 & CE Loss &  P=32  &  P=64 &  P=96 \\
 \hline
 MA (\%)  & 79.43 & 82.42 & \textbf{82.46 } & 82.49\\
\hline
\end{tabular}%
\vspace{-0.3cm}
\caption{Test accuracy of our model with different embedding dimension P on CIFAR-10.} %% of xiushi shui
\label{tab:cifar10_P}
\vspace{-0.6cm}
\end{center}
\end{table}

\noindent\bd{Effects of Removed Loss Values.} Ablations of S removed loss values are reported in Table \ref{tab:cifar10_s}.
One can find the MA has a slight decrease as P increases.
Considering the different requirements for the complete loss value information in different situations, we choose to provide as more loss value information as possible.
We thus set the S equal to 5.
\begin{table}[h]
\vspace{-0.1cm}
\begin{center}
\tabcolsep=0.37cm
\begin{tabular}{c|c|c|c|c}
\hline
 & CE Loss &  S=5  &  S=30 &  S=60 \\
 \hline
 MA (\%)  & 79.43 & \textbf{82.46 } & 82.43 & 82.42\\
\hline
\end{tabular}%
\vspace{-0.30cm} 
\caption{Test accuracy of our model with different S removed loss values on CIFAR-10.} % of
\label{tab:cifar10_s}
\vspace{-0.5cm}
\end{center}
\end{table}

% \FloatBarrier

% \vspace{-0.20cm} 
%------------------------------------------------------------------------
\section{Conclusion}\label{sec:conclusion}

This paper introduces a novel probe-and-allocate training strategy to alleviate class imbalance and corrupted labels in piratically collected data.
Different from prior methods designed to solve either class imbalance or corrupted labels, our method is capable of handling both data biases simultaneously by exploiting informative training loss curve to generate proper sample weights. 
Extensive experiments conducted on synthetic and real-world datasets with various imbalance factors and noise rates well demonstrate the superiority of our method. 
In this paper, however, we only verify our method on classification task, while the proposed approach is generic and could be flexibly applied to training models with biased data on other tasks, which  will be put in future work. 

\section{Acknowledgments}
This work was supported by the National Key Scientific Instrument and Equipment Development Project of China (61527802) and  the Key Laboratory Foundation under Grant TCGZ2020C004 of Science and Technology on Near-Surface Detection Laboratory.

\bibliography{aaai22}

\end{document}